\documentclass[manuscript]{acmart}

\usepackage[utf8]{inputenc} % allow utf-8 input
\usepackage[T1]{fontenc}    % use 8-bit T1 fonts
\usepackage{hyperref}       % hyperlinks
\usepackage{url}            % simple URL typesetting
\usepackage{amsfonts}       % blackboard math symbols
\usepackage{nicefrac}       % compact symbols for 1/2, etc.
\usepackage{microtype}      % microtypography
\usepackage{xcolor}         % colors

\newtheorem{theorem}{Theorem}
\newtheorem{lemma}{Lemma}
\newtheorem{corollary}{Corollary}

\newcommand{\dif}{\mathrm{d}}

\DeclareMathOperator{\barotimes}{\bar{\otimes}}

\setcopyright{acmlicensed}
\copyrightyear{2025}
\acmYear{2025}
\acmDOI{XXXXXXX.XXXXXXX}
\begin{document}

%%
%% The "title" command has an optional parameter,
%% allowing the author to define a "short title" to be used in page headers.
\title{Deriving Transformer Architectures from Implicit Multinomial Regression}

%%
%% The "author" command and its associated commands are used to define
%% the authors and their affiliations.
%% Of note is the shared affiliation of the first two authors, and the
%% "authornote" and "authornotemark" commands
%% used to denote shared contribution to the research.
\author{Jonas A. Actor}
\email{jaactor@sandia.gov}
\orcid{0000-0003-4356-0879}
\author{Anthony Gruber}
\email{adgrube@sandia.gov}
\orcid{0000-0001-7107-5307}
\author{Eric C. Cyr}
\email{eccyr@sandia.gov}
\orcid{0000-0003-3833-9598}
\affiliation{%
  \institution{Sandia National Laboratories}
  \city{Albuquerque}
  \state{New Mexico}
  \country{USA}
}

%%
%% By default, the full list of authors will be used in the page
%% headers. Often, this list is too long, and will overlap
%% other information printed in the page headers. This command allows
%% the author to define a more concise list
%% of authors' names for this purpose.
\renewcommand{\shortauthors}{Actor et al.}

%%
%% The abstract is a short summary of the work to be presented in the
%% article.
\begin{abstract}
From a first-principles standpoint, the attention mechanisms used in state-of-the-art transformer models are opaque. 
% in that 
While attention has been empirically shown to improve model performance, it lacks a rigorous mathematical justification. This short paper establishes a novel connection between attention mechanisms and multinomial regression. Specifically, we show that in a fixed multinomial regression setting, optimizing over latent features yields  solutions that align with the dynamics induced on features by attention blocks. In other words, the evolution of representations through a transformer can be interpreted as a trajectory that recovers the optimal features for classification.
\end{abstract}

%%
%% The code below is generated by the tool at http://dl.acm.org/ccs.cfm.
%% Please copy and paste the code instead of the example below.
%%


%%
%% Keywords. The author(s) should pick words that accurately describe
%% the work being presented. Separate the keywords with commas.

\received{\today}
% \received[revised]{12 March 2009}
% \received[accepted]{5 June 2009}

\maketitle

\section{Introduction}
\label{sec:intro}

Many recent efforts have aimed to provide understanding of modern transformer and large model architectures. 
Among these, mechanistic interpretations \citep{ferrando2024primer, sharkey2025openproblems, olah2020zoom, ameisen2025circuit, cunningham2023sparse, lindsey2025biology} provide \textit{post-hoc} analysis that relate observations of model behavior to specific neuronal or architectural properties, but fail to provide \textit{a priori} analysis or theoretical justification of improved performance. 
From a theoretical standpoint, it remains unclear why certain modern architectures, such as transformers, offer significant improvements over multilayer perceptrons (MLPs).

We hypothesize that the remarkable performance of transformer architectures stems from their implicit execution of feature discovery in conjunction with multinomial regression.
This paper derives how the transformer architecture arises naturally from an optimization problem aimed at recovering the best features for a given multinomial regression task. This connection between transformer blocks and multinomial regression offers a form of \textit{intrinsic interpretability} for the transformer architecture, with insight into designing other attention mechanisms for specialized tasks.

\section{Theoretical Results}
Classification via multinomial logistic regression uses a generalized linear model that minimizes the cross-entropy between its predictions and given class labels. The present setting considers a set of input features (possibly in some latent space) $Z = Z(X) \in \mathbb{R}^{S\times F^i}$, along with categorical labels $C \in [0,1]^{S \times F^o}$ which are one-hot encoded so that $1^\intercal C = \sum_{s=1}^S C_{s,f^o} = 1$ for each feature $f^o \in \{1,\dots,F^o\}$.
Training a model $N(Z,\theta)$ with parameters $\theta\in \mathbb{R}^{F^o\times F^i}$ to fit the labels $C$ involves minimization of the categorical cross-entropy,
\begin{equation}
    \min_{Z,\theta} -\mathbb{E}\left\langle C, \log \sigma_i(N(Z,\theta)) \right\rangle,
\end{equation}
where $\mathbb{E}=\mathbb{E}_{X\sim\rho}$ denotes expectation on $X$ drawn from some distribution $\rho$, and $\sigma_i(A)_{ij} = \left(\sum_{k}\exp{A_{kj}}\right)^{-1}\exp{A_{ij}}$ denotes a softmax applied independently to each sequence\footnote{Note that exp denotes the element-wise exponential and $\sum_{i}\sigma_i(A)_{ij} = 1_j$ for each column $j$.}.
Using the definition of softmax, this expression reduces to the expectation of a difference of convex functions in $N(Z,\theta)$,
\begin{equation}\label{eqn:expanded-ce}
    \min_{Z,\theta} L(Z,\theta) := \min_{Z,\theta} \mathbb{E} \left[ \left\langle 1, \text{LSE}_i( N(Z,\theta))\right\rangle  - \left\langle C, N(Z,\theta) \right\rangle \right],
\end{equation}
where $\text{LSE}_i$ is the log-sum-exponential function applied across the sequence axis, independently for each feature (as in sequence-to-sequence applications of attention), and we have used that $1^\intercal C = 1$.  
We solve this optimization problem via a two stage optimization algorithm that first minimizes over the features $Z$ at each point $X$, with the model $N$ and its parameters $\theta$ fixed. To this end, denote the \emph{point-wise} cross-entropy as 
\begin{equation} \label{eq:pwxe}
\ell(Z,\theta) =
\left\langle 1, \text{LSE}_i( N(Z,\theta))\right\rangle  - \left\langle C, N(Z,\theta) \right\rangle.
\end{equation}
Determining features $Z$ by minimizing this quantity at each point $X$ greedily reduces the global cross-entropy \eqref{eqn:expanded-ce}. 

Consider a linear model $N(Z,\theta) =Z \theta^\intercal$.  
Recall (e.g., \citep{blanchard2021accurately}) that the gradient of LSE is the softmax function; thus,
the gradient of the point-wise cross-entropy with respect to features $Z$ is 
\begin{equation}  
    \partial_Z \ell(Z,\theta) =  1^\intercal \partial_Z \text{LSE}_i( Z\theta^\intercal)  - C \theta 
    = \sigma_i(Z \theta^\intercal)\theta - C \theta  
    = \mathtt{CA}\left(Z, \theta\right) - C \theta,
\end{equation}
where $\mathtt{CA}(Z,\theta) = \sigma(Z\theta^\intercal)\theta$ is the usual cross-attention between $Z$ and $\theta$; see  Appendix~\ref{app:gradients} for a derivation.

% Let $Z_0(X)$ represent an initial guess for the features at each point $X$.
Consider the gradient flow of $\ell(Z,\theta)$ with respect to $Z$ starting from $Z_0(X)$, an initial guess for the features at each point X.
% \footnote{This mapping can be injective to ensure a unique representation for $X$.} 
This flow generates trajectories satisfying the matrix differential equation 
\begin{equation}
    \dot Z(t) = - \partial_Z \ell(Z(t),\theta) = C\theta - \mathtt{CA}(Z(t),\theta), \qquad
    Z(0) = Z_0(X).
\end{equation}
In particular, solutions to this equation are guaranteed to dissipate (point-wise) cross-entropy in continuous time, since
\begin{equation}
    \dot{\ell}(Z,\theta) = \left\langle\partial_Z \ell(Z,\theta),\dot{Z}\right\rangle = -\frac{1}{2} \left\|\partial_Z \ell(Z,\theta)\right\|^2 \leq 0.
\end{equation}
It follows that a trajectory obeying gradient flow simultaneously for all $X$ will minimize the global cross-entropy $L$ over time.
Applying an operator splitting \citep{strang1968construction, mclachlan2002splitting, hairer2006geometric} to this system of ordinary differential equations now yields a system of discrete equations starting from initial condition $Z^{(0)} = Z_0(X)$,
\begin{equation} \label{eq:transformer} \begin{split}
    Z^{(\ell + \frac12 )} &= Z^{(\ell)} - \mathtt{CA}\left(Z, \theta^{(\ell)} \right) \\
    Z^{(\ell+1)} &= Z^{(\ell + \frac12)} + C \theta^{(\ell+\frac12)}.
\end{split}
\end{equation}
Remarkably, this mirrors the transformer block in \citep{vaswani2017attention} using cross-attention. The first line of Equation \eqref{eq:transformer} performs cross-attention with a residual connection, while the second line performs a linear mapping of cross-attention's keys/values (i.e., multiplication by $C$), followed by another residual connection. Note that the keys/values are \textit{not} a function of the queries $Z$ in this case.
Treating Equations~\eqref{eq:transformer} as layer-wise propagation of feature information, the parameters $\theta^{(\ell)}$ and $\theta^{(\ell+\frac12)}$ can then be optimized via gradient descent (or any other optimizer of choice). 

Since the use of a linear model $N(Z,\theta) = Z\theta^\intercal$ for multinomial logistic regression has connections to cross-attention, it is interesting to consider a quadratic model $N(Z,\theta) = Z\theta Z^\intercal$ where $\theta\in\mathbb{R}^{F^i\times F^i}$ is square.  For simplicity, suppose that $\theta=\theta^\intercal$ is symmetric and positive definite, so that $\theta = \phi \phi^\intercal$ for some lower-triangular square root $\phi\in\mathbb{R}^{F^i\times F^i}$.
Note that in this case, the class target $C\in\mathbb{R}^{S\times S}$ is now a square matrix. Using Equation~\eqref{eq:pwxe} and repeating the steps above (see Corollary~\ref{lem:quad}  in Appendix~\ref{app:gradients} for a derivation), the gradient of the point-wise cross entropy with respect to $Z$ becomes
\begin{equation} \begin{split}
    \partial_Z \ell(Z,\theta) &=  2 \sigma_i(Z\theta Z^\intercal)Z\theta - (C+C^\intercal) Z\theta  \\
    &= 2 \, \mathtt{SA}(Z \phi) \phi^\intercal  - (C+C^\intercal) Z \theta,
     \end{split}
\end{equation}
where $\mathtt{SA}(X) = \sigma_i(XX^\intercal)X$ is the usual self-attention.  
% For a derivation of this gradient, see Corollary \ref{lem:quad} in the Appendix.
Applying the same gradient flow of point-wise cross-entropy, followed by an operator splitting, we arrive at the corresponding discrete equations, with initial conditions $Z^{(0)} = Z_0(X)$,
\begin{equation} \label{eq:self_transformer} \begin{split}
    Z^{(\ell + \frac12 )} &= Z^{(\ell)} - 2\, \mathtt{SA}\left(Z\phi^{(\ell)}\right) {\phi^{(\ell)}}^\intercal  \\
    Z^{(\ell+1)} &= Z^{(\ell + \frac12)} + (C + C^\intercal) Z^{(\ell + \frac12)} \theta^{(\ell+\frac12)}.
\end{split}
\end{equation}
Again, this structure closely mirrors that of a transformer block with self-attention: the first step performs self-attention followed by a residual connection, while the second step applies a linear transformation.
% Returning to the cross-entropy, note that choosing features which minimize the \emph{point-wise} cross-entropy for a given $\theta$ is sufficient to ensure minimization over $Z$ in Equation~\eqref{eqn:expanded-ce}.  
Minimizing over $\theta$ then becomes the training problem typically solved using a first-order optimizer of choice.

\section{Numerical Experiments}
This architectural relationship between attention and classification is tested with the following numerical experiment. We first trained a linear classifier on Fashion MNIST using data corrupted by Gaussian noise. These noisy images formed the input $Z^{(0)}$ which was passed through 5 iterations of the attention block in Equation \eqref{eq:transformer}, where the parameters $\theta^{(\ell)} = \theta^{(\ell+\frac12)} = \theta$ are the coefficients of the optimal linear classifier.  The results of this were then compared to the same attentional update with input $X$ comprised of the original images without corruption.  More details on this setup can be found in Appendix \ref{app:experimental_details}.

\begin{figure}[ht]
\centering
\includegraphics[width=0.95\textwidth]{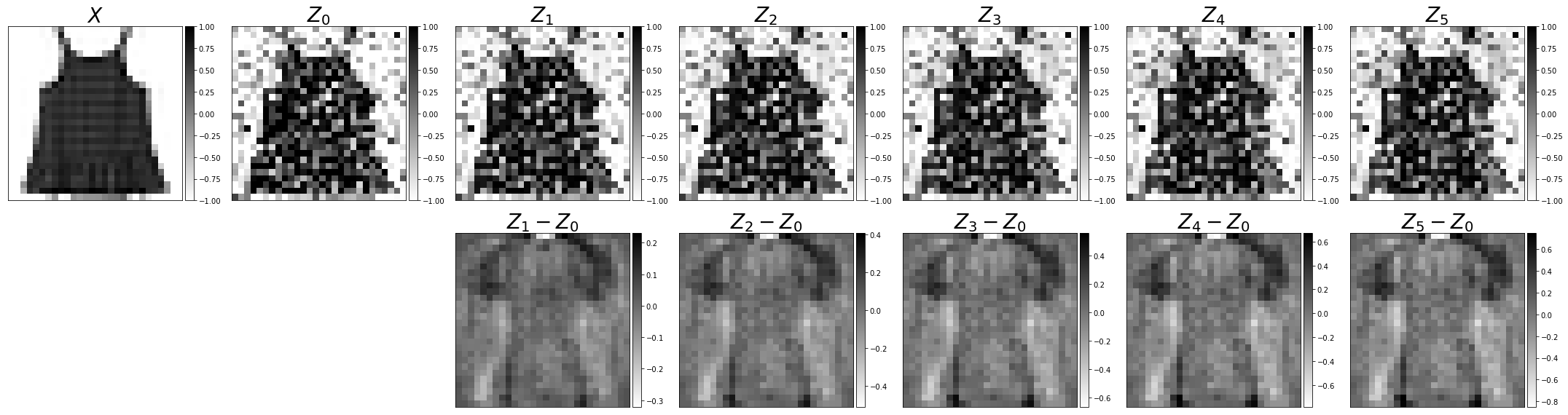}
% \caption{Evolution of $Z^{(\ell)}$ when iterating through a transformer block, from original image $X$, initial (noisy) value $Z^{(0)}$, and subsequent iterates $Z^{(\ell)}$. Second line tracks the difference between $Z^{(0)}$ and $Z^{(\ell)}$ (with different colorbars for each image).  Note the substantial lightening around the collar, and darkening above the sleeves, in each iterate. \label{fig:example} }
\caption{Evolution of a sample $Z^{(\ell)}$ when iterating through a transformer block. First line shows the original image $X$, initial (noisy) value $Z^{(0)}$, and subsequent iterates $Z^{(\ell)}$. Second line tracks the difference between $Z^{(0)}$ and $Z^{(\ell)}$, with substantial lightening around the collar, and darkening above the sleeves, in each iterate. \label{fig:example}}
\Description{Figure showing the evolution of a sample $Z^{(\ell)}$ when iterating through a transformer block. First line shows the original image $X$, initial (noisy) value $Z^{(0)}$, and subsequent iterates $Z^{(\ell)}$. Second line tracks the difference between $Z^{(0)}$ and $Z^{(\ell)}$, with substantial lightening around the collar, and darkening above the sleeves, in each iterate. }
\end{figure}

Table \ref{table:acc} shows substantial improvement in classification of both the noisy features $Z^{(0)}$ and the original images $X$ with each pass.
\begin{table}
  \caption{Accuracy per iteration through attention block, starting from both clean and noisy data. Even a single pass through Equation \eqref{eq:transformer} produces a substantial improvement in accuracy.}
  \label{table:acc}
  \begin{tabular}{lrrrrrr}
    \toprule
    Iteration & 0 & 1 & 2 & 3 & 4 & 5 \\
    \midrule
    Clean & 0.8424 & 0.9788 &	0.9963 &	0.9992 &	0.9998 &	0.9999 \\
    Noisy &	0.8139 & 0.9835 &	0.9978 &	0.9999 &	1.0000 &	1.0000 \\
  \bottomrule
\end{tabular}
\end{table}
The corrections applied at each iteration appear to emphasize archetypal features for each class. As an illustration, Figure \ref{fig:example} depicts the trajectory of an image through iterations of Equation \eqref{eq:transformer}; while the initial classifier applied to $Z^{(0)}$ mistakes this image for a bag, the same classifier correctly identifies the final $Z^{(5)}$ as a t-shirt/top. 
We note that there is no training of the attention block in this experiment; a simple forward pass significantly improves classification accuracy.

\section{Conclusions and Limitations}
The derivation shows that the structure of the \textit{transformer architecture itself} implicitly propagates features through gradient flow of the cross-entropy, connecting nicely to standard multinomial regression.  Changes to the original cross-entropy minimization problem, such as adding constraints, or alterations to the splitting method, provide multiple avenues for exploration and to develop attention mechanisms for specialized tasks.

The primary limitation of this analysis lies in the selected model form for $N(Z,\theta)$, which is a restricted form of the MLP layer typically present in transformer blocks. We also do not account for the effects of normalization layers, which play a significant role in practical transformer architectures. Notably, normalization shares similarities with whitening, as used in, e.g., independent component analysis in place of multinomial regression. Alternatively, normalization can be viewed as constraining features to the sphere, effectively imposing a  directional constraint on the classification problem. 

Moreover, this analysis assumes the true class labels $C$ are available during training, and it is unclear how this will extend to, e.g., self- or semi-supervised training approaches.  An interesting direction for future work is to relax this assumption by introducing an oracle (possibly learned) that provides the labels, potentially integrating with the attention mechanisms within a self-supervised framework.
An oracle emulated via an MLP would bridge the gap between our theoretical formulation and modern state-of-the-art architectures.

Finally, while we offer an interpretation of the underlying mechanics of attention, modern architectures are highly complex and often incorporate additional components beyond the scope of this analysis. 
Extension to e.g., multi-head attention mechanisms should be straightforward.  Ultimately, we hope that the ideas presented here will prove fruitful for understanding attention through the lens of gradient-based dynamics and inspire additional work in this area.

\begin{acks}
 Sandia National Laboratories is a multi-mission laboratory managed and operated by National Technology \& Engineering Solutions of Sandia, LLC (NTESS), a wholly owned subsidiary of Honeywell International Inc., for the U.S. Department of Energy’s National Nuclear Security Administration (DOE/NNSA) under contract DE-NA0003525. This written work is authored by an employee of NTESS. The employee, not NTESS, owns the right, title and interest in and to the written work and is responsible for its contents. Any subjective views or opinions that might be expressed in the written work do not necessarily represent the views of the U.S. Government. The publisher acknowledges that the U.S. Government retains a non-exclusive, paid-up, irrevocable, world-wide license to publish or reproduce the published form of this written work or allow others to do so, for U.S. Government purposes. The DOE will provide public access to results of federally sponsored research in accordance with the DOE Public Access Plan. SAND\#2025-10835O. 
\end{acks}

{

\bibliographystyle{ACM-Reference-Format}
\bibliography{ailet}

}

%%%%%%%%%%%%%%%%%%%%%%%%%%%%%%%%%%%%%%%%%%%%%%%%%%%%%%%%%%%%

\appendix

\section{Supplemental Derivations for Gradient Calculations}
Here we record some computations necessary for fully appreciating the results in the main body.

% \subsection{Gradients Involving Cross-Entropy and the Log-Sum-Exp Function}
\label{app:gradients}

The gradient of the cross-entropy $L$ requires differentiating the vector-valued function $\text{LSE}_i(Z\theta^\intercal)$.  This can be described succinctly in terms of the operation $A\barotimes B$ on matrices $A,B$, which denotes the order-3 tensor with components $(A\barotimes B)_{ijk} = A_{ij}B_{ik}$.  Note that the row dimensions of $A,B$ must match in order for this diagonal extraction to make sense.

\begin{theorem}\label{thm:easygrad}
    The $Z$-derivative of log-sum-exp applied to the linear model $Z\theta^\intercal$ satisfies $\partial_Z \mathrm{LSE}_i(Z\theta^\intercal) = \sigma_i(Z\theta^\intercal)^\intercal\barotimes\theta$.
\end{theorem}
\begin{proof}
    Since $\text{LSE}_i(X\theta^\intercal)\in\mathbb{R}^s$ is vector valued, so is its exterior derivative in $Z$ (currying $\theta$), which computes to 
    \begin{align*}
    \dif_Z\text{LSE}_i(Z\theta^\intercal) &= \dif_Z\log\left(1^\intercal \exp(Z\theta^\intercal)\right) = 1^\intercal \frac{\exp(Z\theta^\intercal)\odot \dif Z\,\theta^\intercal}{1^\intercal\exp(Z\theta^\intercal)} = 1^\intercal \left(\sigma_i(Z\theta^\intercal)\odot \dif Z\,\theta^\intercal \right) \\
    &= \left(\sigma_i(Z\theta^\intercal)^\intercal\barotimes\theta\right) : \dif Z, 
    \end{align*}
    where $\odot$ denotes element-wise multiplication and the last equality has used that
    \begin{align*}
    \sum_{i,k} \sigma_i(Z\theta^\intercal)_{ij} \dif Z_{ik}\theta_{jk} = \sum_{i,k}\left[\sigma_i(Z\theta^\intercal)_{ij}\theta_{jk}\right] \dif Z_{ik} = \sum_{j,k} \left[\sigma_i(Z\theta^\intercal)^\intercal\barotimes\theta\right]_{ijk} \dif Z_{jk}.
    \end{align*}
    Since the exterior derivative $\dif f(Z)\in\mathbb{R}^s$ of a function $f:\mathbb{R}^{s\times f}\to\mathbb{R}^s$ has the expression $\nabla f(Z): \dif Z$ in terms of the gradient $\nabla f(Z)\in\mathbb{R}^{s\times s\times f}$ at $Z$, the conclusion follows.
\end{proof}

Differentiating the term $\mathbb{E}\left\langle 1,\text{LSE}_i(X\theta^\intercal)\right\rangle$ appearing in the cross-entropy $L$ is now as simple as observing that 
% \[ \partial_Z \left\langle 1,\text{LSE}_j(Z\theta^\intercal)\right\rangle = 1^\intercal\left(I\barotimes\sigma_j(Z\theta^\intercal)Z\right) = \sigma(Z\theta^\intercal)Z = \mathtt{CA}(Z,\theta).\]
\[ \partial_Z \left\langle 1,\text{LSE}_i(Z\theta^\intercal)\right\rangle = 1^\intercal\left(\sigma_i(Z\theta^\intercal)^\intercal\barotimes\theta\right) = \sigma(Z\theta^\intercal)\theta = \mathtt{CA}(Z,\theta).\]
Alternatively, there is the following direct calculation for the gradient of the cross-entropy which short-circuits the more elaborate computation in Theorem~\ref{thm:easygrad}.  This will be computed for a general differentiable model $N(Z,\theta)$ and relies on the following simple Lemma.
\begin{lemma}\label{lem:gradlse}
    The gradient of the scalar function $f(X) = \mathrm{tr}\left(\mathrm{LSE}_i(X)\right) = \langle 1,\mathrm{LSE}_i(X)\rangle$ is given by $\nabla f(X) = \sigma_i(X)$.
\end{lemma}
\begin{proof}
    Noting that the exterior derivative of $\text{LSE}_i$ is 
    \[\dif \,\text{LSE}_i(X) = \dif \log\left(1^\intercal\exp(X)\right) = \frac{1^\intercal(\exp(X)\odot \dif X)}{1^\intercal\exp(X)} =  1^\intercal\left(\sigma_i(X)\odot \dif X\right), \]
    the exterior derivative of $f$ computes to
    \begin{align*}
        \dif f(X) &= \left\langle 1, \dif \,\text{LSE}_i(X)\right\rangle = \left\langle 1 , 1^\intercal\left(\sigma_i(X)\odot \dif X\right)\right\rangle = \left\langle 11^\intercal, \sigma_i(X)\odot \dif X\right\rangle \\
        &= \left\langle \sigma_i(X), \dif X\right\rangle = \left\langle \nabla f(X), \dif X\right\rangle,
    \end{align*}
    since $\langle A, B \odot C\rangle = \langle A\odot B, C\rangle$, directly yielding the desired gradient.
\end{proof}

This provides a way to compute the $Z$-gradient of the cross-entropy for any desired model $N(Z,\theta)$.
\begin{theorem}\label{thm:gengrad}
    The $Z$-gradient of the cross entropy $L(Z,\theta)$ satisfies $\partial_ZL(Z,\theta) = \mathbb{E}\left[\left(\sigma_i(N(Z,\theta)) - C\right): \partial_ZN(Z,\theta)\right]$ in terms of the model $Z$-gradient $\partial_ZN(Z,\theta)$.
\end{theorem}
\begin{proof}
    Recalling that $L(Z,\theta) = \mathbb{E} \left[ \left\langle 1, \text{LSE}_i\left(N(Z,\theta)\right)\right\rangle  - \left\langle C, N(Z,\theta) \right\rangle \right],$ we work term-by-term.  The second term under the expectation has the $Z$-exterior derivative
    \[ \dif_Z\langle C, N(Z,\theta)\rangle = \langle C, \partial_Z N(Z,\theta): \dif Z\rangle = \langle C:\partial_ZN(Z,\theta), \dif Z\rangle,\]
    implying that $\partial_Z \langle C, N(Z,\theta)\rangle = C:\partial_ZN(Z,\theta)$.  Computing the same for the first term in view of Lemma~\ref{lem:gradlse} yields 
    \begin{align*}
        \dif_Z \left\langle 1, \text{LSE}_i\left(N(Z,\theta)\right)\right\rangle &= \left\langle \sigma_i(N(X,\theta)), \dif N(Z,\theta)\right\rangle = \left\langle \sigma_i(N(X,\theta)), \partial_ZN(Z,\theta):\dif Z\right\rangle \\
        &=  \left\langle \sigma_i(N(X,\theta)):\partial_ZN(Z,\theta), \dif Z\right\rangle,
    \end{align*}
    establishing the gradient $\partial_Z\mathrm{LSE}_i\left(N(Z,\theta)\right) = \sigma_i(N(X,\theta)):\partial_ZN(Z,\theta).$ Combining this with the vanishing of the operator commutator $[\partial_Z, \mathbb{E}]=0$ directly yields the conclusion.
\end{proof}

Theorem~\ref{thm:gengrad} now provides a recipe for computing derivatives of the cross-entropy under the linear and quadratic models considered in this paper.

\begin{corollary} \label{lem:linear}
    The $Z$-gradient of the cross entropy $L(Z,\theta)$ for a linear model $N(Z,\theta) = Z\theta^\intercal$ satisfies $\partial_ZL(Z,\theta) = \mathbb{E}\left[ \mathtt{CA}\left(Z, \theta\right) - C \theta \right].$
\end{corollary}
\begin{proof}
    Observe that the $Z$-gradient of $N(Z,\theta) = Z\theta^\intercal$ is given by $\left[\partial_Z(Z\theta^\intercal)\right]_{ijkl} = \delta_{ik}\theta_{jl}$ since 
    \[\dif_ZN(Z,\theta)_{ij} = (\dif Z\,\theta^\intercal)_{ij} = \sum_k \dif Z_{ik}\theta_{jk} = \sum_{k,l} \delta_{il}\theta_{jk}\dif Z_{lk}.\]
    It follows that $M : \partial_Z(Z\theta^\intercal) = M\theta$ for any matrix $M$, and therefore Theorem~\ref{thm:gengrad} directly implies
    \[\partial_ZL(Z,\theta) = \mathbb{E}\left[\left(\sigma_i(Z\theta^\intercal) - C\right): \partial_Z(Z\theta^\intercal)\right] = \mathbb{E}\left[\sigma_i(Z\theta^\intercal)\theta - C\theta\right],\]
where $\mathtt{CA}(Z,\theta) = \sigma_i(Z\theta^\intercal)\theta$ as claimed.
\end{proof}

\begin{corollary} \label{lem:quad}
    The $Z$-gradient of the cross entropy $L(Z,\theta)$ for a quadratic model $N(Z,\theta) = Z\theta Z^\intercal$ with symmetric $\theta$ satisfies $\partial_ZL(Z,\theta) = \mathbb{E}\left[ 2 \sigma_i(Z\theta Z^\intercal)Z\theta - (C+C^\intercal) Z\theta  \right].$
\end{corollary} 
\begin{proof}
    Observe that the $Z$-gradient of $N(Z,\theta) = Z\theta Z^\intercal$ is given by $\left[\partial_Z(Z\theta Z^\intercal)\right]_{ijkl} = \delta_{ik}(Z\theta^\intercal)_{jl} + \delta_{jk}(Z\theta)_{il}$ since 
    \begin{align*}
        \dif_ZN(Z,\theta)_{ij} &= (\dif Z\,\theta Z^\intercal)_{ij} + (Z\theta\,\dif Z^\intercal)_{ij} = \sum_{k} \dif Z_{ik}(\theta Z^\intercal)_{kj} + \sum_{k} (Z\theta)_{ik}\dif Z_{jk} \\
        &= \sum_{k,l} \left(\delta_{il}(\theta Z^\intercal)_{kj} + \delta_{jl}(Z\theta)_{ik}\right)\dif Z_{lk} = \sum_{k,l}\left(\delta_{ik}(Z\theta^\intercal)_{jl} + \delta_{jk}(Z\theta)_{il}\right)\dif Z_{kl}.
    \end{align*}
    It follows that $M : \partial_Z(Z\theta Z^\intercal) = MZ\theta^\intercal + M^\intercal Z \theta$ for any matrix $M$, and therefore applying Theorem~\ref{thm:gengrad} yields
    \begin{align*}
        \partial_ZL(Z,\theta) &= \mathbb{E}\left[\left(\sigma_i(Z\theta Z^\intercal) - C\right): \partial_Z(Z\theta Z^\intercal)\right] \\
        &= \mathbb{E}\left[ 2 \sigma_i(Z\theta Z^\intercal)Z\theta^\intercal  - CZ\theta^\intercal - C^\intercal Z \theta  \right],
    \end{align*}
    which becomes the stated gradient when $\theta=\theta^\intercal$ since $\sigma_i(Z\theta Z^\intercal)^\intercal = \sigma_j(Z\theta Z^\intercal) = \sigma_i(Z\theta Z^\intercal)$.
\end{proof}

\section{Supplemental Experiment Details}
\label{app:experimental_details}
In this section, we describe additional details for the numerical experiment in the main body.

The corrupted Fashion MNIST dataset is split via a 80\%-20\% train-validation split. For each image, Gaussian noise is then generated anew for each instance when data is loaded. The added Gaussian noise is entrywise independent and identically distributed with mean $0$ and standard deviation of $\frac13$. 

The linear classifier is trained to minimize the cross-entropy loss via an Adam optimizer with a learning rate of $10^{-3}$ and a batch size of 1024 for 100 epochs. We use an Nvidia a100 GPU with 40GB RAM deploying 32 workers with a prefetch factor of 4 for data loading, to expedite the time required for training, which takes less than three minutes to train. A linear classifier on the clean data using e.g. scikit-learn returns an accuracy of around 85\%, so the model performance is suitable compared to other off-the-shelf methods for building a linear classifier.

\end{document}